\documentclass[conference]{IEEEtran}
\IEEEoverridecommandlockouts
\usepackage{cite}
\usepackage{amsmath,amssymb,amsfonts}
\usepackage{algorithmic}
\usepackage{graphicx}
\usepackage{textcomp}
\usepackage{xcolor}
\usepackage{booktabs} 
\usepackage{multirow}
\usepackage{threeparttable}
\usepackage{verbatim}
\usepackage{multicol}
\usepackage{algorithm}
\usepackage{underscore}
\usepackage{amsmath}
\usepackage{bm}
\usepackage{afterpage}
\def\BibTeX{{\rm B\kern-.05em{\sc i\kern-.025em b}\kern-.08em
    T\kern-.1667em\lower.7ex\hbox{E}\kern-.125emX}}
\begin{document}

\title{Robustifying DARTS by Eliminating Information\\Bypass Leakage via Explicit Sparse Regularization
}

\author{\IEEEauthorblockN{Jiuling Zhang}
\IEEEauthorblockA{University of Chinese Academy of Sciences\\
Beijing, China\\
Email: zhangjiuling19@mails.ucas.ac.cn
}
\and
\IEEEauthorblockN{Zhiming Ding\thanks{\IEEEauthorrefmark{4} Corresponding author.}\IEEEauthorrefmark{4}}\footnote{Corresponding author}
\IEEEauthorblockA{Institute of Software Chinese Academy of Sciences\\
Beijing, China\\
Email: zhiming@iscas.ac.cn}
}

\maketitle
\begin{abstract}
Differentiable architecture search (DARTS) is a promising end to end NAS method which directly optimizes the architecture parameters through general gradient descent. However, DARTS is brittle to the catastrophic failure incurred by the skip connection in the search space. Recent studies also cast doubt on the basic underlying hypotheses of DARTS which are argued to be inherently prone to the performance discrepancy between the continuous-relaxed supernet in the training phase and the discretized finalnet in the evaluation phase. We figure out that the robustness problem and the skepticism can both be explained by the information bypass leakage during the training of the supernet. This naturally highlights the vital role of the sparsity of architecture parameters in the training phase which has not been well developed in the past. We thus propose a novel sparse-regularized approximation and an efficient mixed-sparsity training scheme to robustify DARTS by eliminating the information bypass leakage. We subsequently conduct extensive experiments on multiple search spaces to demonstrate the effectiveness of our method.
\end{abstract}


%
\IEEEpeerreviewmaketitle

\section{Introduction}
DARTS constructs a cell-based search space with $N$ nodes $X = \{ {x^1},{x^2},...,{x^N}\} $ and $E$ compound edges $G = \{ g_{_{1,2}}^1,g_{_{1,3}}^2,...,g_{_{N - 1,N}}^E\} $ where every node represents feature maps and the compound edge ${g_{i,j}}$ subsumes all operation candidates to express the transformation from node $i$ to $j$. They explicitly parameterize the neural architecture by associating ${g_{i,j}}$ with three attributes: candidate operation set ${O_{i,j}} = \left\{ {o_{i,j}^1,o_{i,j}^2,...,o_{i,j}^M} \right\}$, corresponding operation parameter set ${A_{i,j}} = \left\{ {\alpha _{i,j}^1,\alpha _{i,j}^2,...,\alpha _{i,j}^M} \right\}$, probability distribution of the parameters ${\bm{a}_{i,j}} = softmax ({A_{i,j}})$. Every intermediate node is connected to all its predecessor through an edge ${x_j} = {g_{i,j}}({x_i})$ where ${g_{i,j}}({x_i}) =  < {\bm{a}_{i,j}},{O_{i,j}}({x_i}) > $. In general, an unified set of operation candidates $O = \left\{ {o^1,o^2,...,o^M} \right\}$ is defined for all edges in the cell. Supernet refers to the network that encodes all architecture candidates. 

Let $categorical(O)$ denotes the categorical selection over all operations within $O$. DARTS overall hinges on two important hypotheses. First, \textit{\textbf{DARTS hypothesizes that the categorical choice of the operations can be approximated through continuous relaxation of the architecture parameters $\bm{\alpha}$}} as shown in Eq.(1) and Eq.(2).
\begin{equation}
categorical(O) \approx softmax (A) = \frac{{\exp (\alpha _{i,j}^o)}}{{\sum\nolimits_{o' \in O} {\exp (\alpha _{i,j}^{o'})} }}
\end{equation}
\begin{equation}
\sum\limits_{m = 1}^M {categorical(O)} o_{i,j}^m({x_i}) \approx \sum\limits_{m = 1}^M {softmax (A)} o_{i,j}^m({x_i})
\end{equation}
In this way, both architecture parameters and operation weights in the supernet can be optimized continuously. DARTS is framed as solving a bilevel optimization objective depicted in Eq.(3).
\begin{equation}
\mathop {\min }\limits_\alpha  {L_{val}}({\omega ^ * }(\alpha ),\alpha ){\rm{\ \ s}}{\rm{.t}}{\rm{.\ }}{\omega ^ * }(\alpha ) = \arg \mathop {\min }\limits_\omega  {L_{train}}(\omega ,\alpha )
\end{equation}
where architecture parameters $\alpha$ and operation weights $\omega$ are alternatively optimized on validation set and training set respectively which enables a highly efficient architecture search through generic gradient optimizer in an end-to-end manner. Henceforth, we abbreviate operation weights as weights and architecture parameters as parameters. 

In summary, hypothesis 1 aims to jointly optimize weights and parameters by gradient descent through a differentiable categorical approximation depicted in Eq.(1) within a bilevel training scheme depicted in Eq.(3).

The second hypothesis is that \textit{\textbf{DARTS expects the gradient-based optimization automatically assigns the largest parameter value to the most important operation within each compound edge.}} If this is the case, we can easily derive the resulting architecture (finalnet) by selecting the operation associated with the largest parameter as shown in Eq.(4) at the end of training which can also be regarded as the $categorical(O)$ approximation in the post-pruning discretization step. 
\begin{equation}
categorical(O) \approx  \arg {\max _{o \in O}} (\bm{a}){\rm{\ \ for\ }}\bm{a}=softmax (A)
\end{equation}
We refer to their paper \cite{liu2018darts} for more details about DARTS.

The training recipe of DARTS is essentially based on the weight-sharing paradigm which has been widely questioned for ranking incompetence and poor standalone surrogate performance \cite{li2021geometry,pourchot2020share,zhang2020deeper,yu2019evaluating}. Problem becomes even worse when the search space accommodates the skip connection. Empirically, DARTS sometimes suffers catastrophic failure when the skip connection becomes gradually dominant during optimization \cite{arber2020understanding,dong2019bench,liang2019darts+}. Most of the previous studies are focused on working out the performance collapse caused by the skip connection to enhance the robustness of DARTS. Both DARTS+~\cite{liang2019darts+} and \cite{wang2020rethinking} theoretically explain this collapse based on a view of minimizing the variances of the feature maps. Assigning stronger strength to the skip connection obviously makes the network easier to be trained and brings lower variances to the deeper layers which as a whole is referred to the unfair advantage in an exclusive competition in Fair DARTS \cite{chu2020fair}. Wang et al. \cite{wang2020rethinking} also gives some practical evidences that the parameter values do not really manifest the importances of corresponding operations.

\begin{figure}[ht]
\begin{center}
\includegraphics[width=\columnwidth]{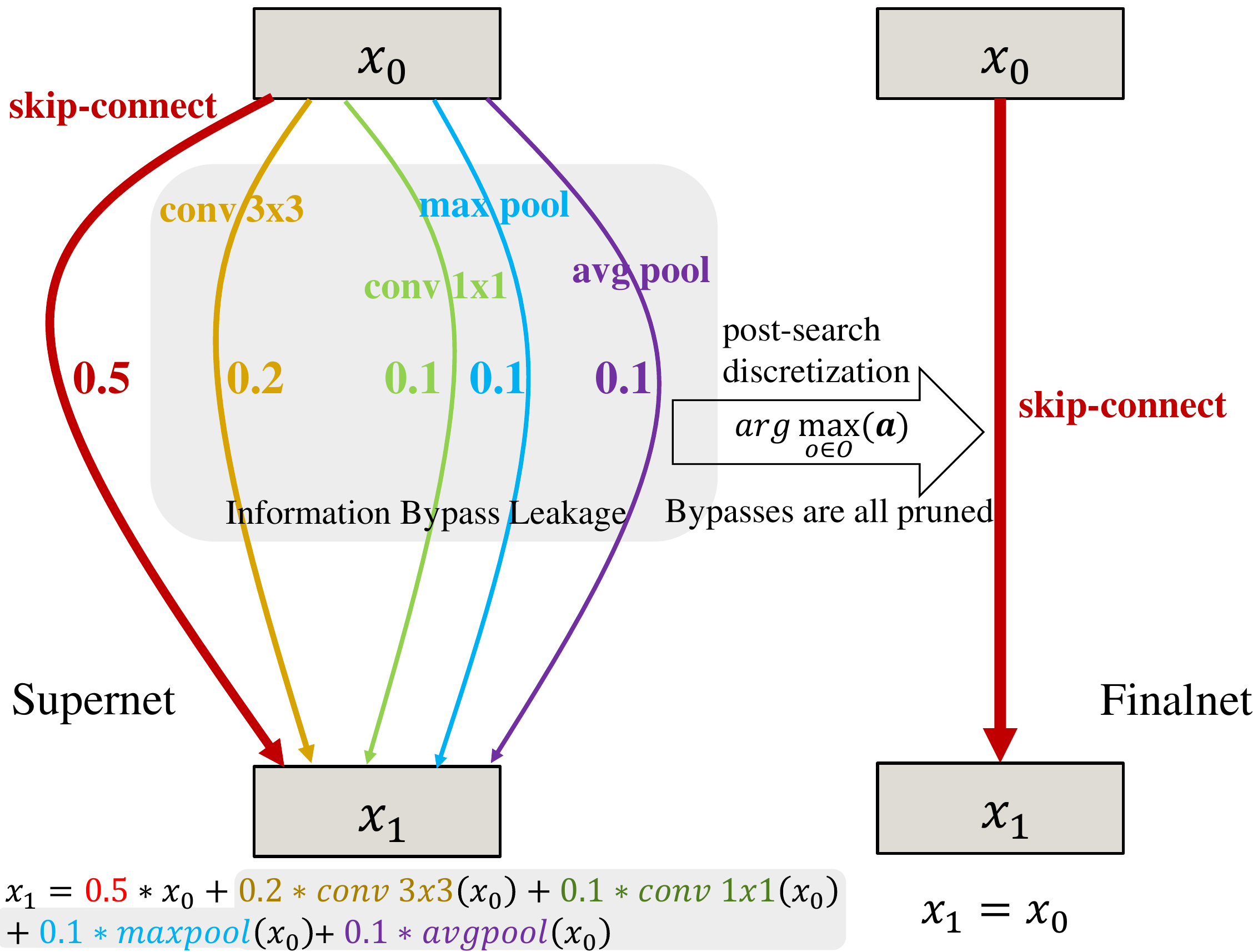}
\caption{A conceptual visualization of the information bypass leakage (shaded by grey rectangle) in the training phase of DARTS.}
\label{fig1}
\end{center}
\end{figure}

Continuous relaxation depicted in Eq.(1) allows all predictive information provided by the subordinate operations ($O\setminus\arg\mathop {\max }\limits_o (\bm{a})$) still be transmitted through supernet in the training phase. Figure~\ref{fig1} schematically illustrates the feature maps transformation from node $x_0$ to $x_1$ where the subordinate operations are covered by faded-grey rectangle. We can clearly see in Figure~\ref{fig1} that even the skip connection dominates the compound edge via the largest parameter distribution component (0.5), the softmax relaxation causes remaining operation outputs still be transmitted through bypasses within faded-grey rectangle. We call this kind of \textit{\textbf{information transmission through supernet due to the differentiable relaxation as the information bypass leakage in DARTS}}. This leakage in turn encourages the optimization to lean more towards the advantage of the skip connection because the network can keep the predictive information unchanged by simply scaling the output of all other operations while comfortably enjoying easier training advantage obtained through the dominant skip connection. Supernet then tries to learn the parameter-weighted optimal fit between different operations. Smooth DARTS \cite{chen2020stabilizing} on the other hand, tries to robustify DARTS by adding noise to the parameters in the training phase. Their strategy forces the supernet to be insensitive of the perturbation of parameters which can be regarded as the interference against the coadaptation of weights.

Another research branch explains this problem from a sparsity aspect, that is, hypothesis 2 brings about some mismatches between supernet and finalnet due to the insufficient sparsity of parameters during training. When we say sparse parameters in DARTS, we refer to the sparsity of the softmax output distribution $\bm{a}=softmax(A)$. As shown on the right side of Figure~\ref{fig1}, bypasses are all pruned in the post-search discretization step. The predictive information transmitted through bypasses in the training phase is no longer exist in the finalnet in the evaluation phase which ultimately leads to the performance discrepancy between training and evaluation. GAEA \cite{li2021geometry} emphasizes that obtaining sparse final architecture parameters is critical for good performance, both for the mixture relaxation, where it alleviates the effect of overfitting, and for the stochastic relaxation, where it reduces noise when sampling architectures. Based on this view, GAEA explicitly guides sparse training by increasing the parameter learning rate (0.0003$\to$0.1) and employing mirror decent. However, increasing the learning rate is incapable to make the parameters reach the sufficient level of sparsity to suppress information bypass leakage due to the gradient saturation of softmax. In addition, the mirror-decent-based GAEA-Bilevel is empirically more prone to the performance collapse than the original DARTS. Fair DARTS \cite{chu2020fair} breaks the unique advantage of the skip connection by replacing the softmax with sigmoid. This way, Fair DARTS completely abandons the exclusivity of the operation selection and turns to explicitly learn the best parameter-weighted fitness across multiple operations.

Additionally, FBNet \cite{wu2019fbnet}, SNAS \cite{xie2018snas}, GDAS \cite{dong2019searching} absolutely discard the hypothesis 1 and utilize the gumbel softmax to mimic the categorical operation selection. GDAS samples a sub-graph from the whole DAG by sampling one feature in a differentiable way between every two nodes to achieve absolute sparsity in the training phase. However, GDAS suffers premature convergence to sub-optimal \cite{chen2020stabilizing}. The discretization of the gumbel softmax manifests search instability \cite{chen2021drnas,zela2019bench}. Empirically, GDAS needs about 3$\times$ time overhead compared with DARTS-V1 on NAS-BENCH-201 \cite{dong2019bench}. In this paper, we propose to resort to small temperature rather than the reparameterization trick to facilitate parameter distribution much sparser than any previous research to suppress the information bypass leakage which in turn both eliminates catastrophic failure and alleviates  performance discrepancy in DARTS. Our contributions can be summarized as:
\begin{itemize}
\setlength{\itemsep}{0pt}
\setlength{\parsep}{0pt}
\setlength{\parskip}{0pt}
\item We reformulate the objective of DARTS to give a new sparse-regularized differentiable approximation of the categorical operation selection;
\item We propose an efficient training scheme to gradually increase the strength of the sparse regularization and guide parameters progressively converge to the sparse solution with negligible additional overhead;
\item We conduct extensive experiments on multiple datasets and search spaces to validate the efficacy of our method.
\end{itemize}

\section{Sparse Regularization}
We first neglect two hypotheses and give the ideal objective of DARTS in Eq.(5).
\begin{equation}
\mathop {\min }\limits_c  {L_{val}}(\omega _c^ * (c),c)
\end{equation}
\begin{equation}
{\rm{s}}{\rm{.t}}{\rm{.\ }}\left\{ {\begin{array}{*{20}{c}}
\begin{aligned}
c &= categorical(O)\\
\omega _c^ * (c) &= \arg \mathop {\min }\limits_\omega  {L_{train}}(\omega ,c)
\end{aligned}
\end{array}} \right.
\end{equation}
One crucial flaw of Eq.(5) is the non-differentiable $categorical(O)$. DARTS continuously relaxes $categorical(O)$ to $softmax(A)$ based on hypothesis 1 as shown in Eq.(7).
\begin{equation}
\mathop {\min }\limits_\alpha {L_{val}}(\omega _\alpha^ * (\bm{a}),\bm{a})
\end{equation}
\begin{equation}
{\rm{s}}{\rm{.t}}{\rm{. }}\left\{ {\begin{array}{*{20}{c}}
\begin{aligned}
\bm{a} &= softmax (A)\\
\omega _\alpha^ * (\bm{a}) &= \arg \mathop {\min }\limits_\omega  {L_{train}}(\omega ,\bm{a})
\end{aligned}
\end{array}} \right.
\end{equation}
For sparse training, intuitively, one straightforward idea is to accomplish a sparse approximation by employing a small temperature value $t_{sp}$ depicted in Eq.(9).
\begin{equation}
\begin{array}{l}
\begin{aligned}
categorical(O) &\approx softmax (\frac{A}{t_{sp}})\\
 &= \frac{{\exp ({\raise0.7ex\hbox{${\alpha _{i,j}^o}$} \!\mathord{\left/
 {\vphantom {{\alpha _{i,j}^o} t}}\right.\kern-\nulldelimiterspace}
\!\lower0.7ex\hbox{$t_{sp}$}})}}{{\sum\nolimits_{o' \in O} {\exp ({\raise0.7ex\hbox{${\alpha _{i,j}^{o'}}$} \!\mathord{\left/
 {\vphantom {{\alpha _{i,j}^{o'}} t}}\right.\kern-\nulldelimiterspace}
\!\lower0.7ex\hbox{$t_{sp}$}})} }}
\end{aligned}
\end{array}
\end{equation}
where we discard the reparameterization trick and resort to the temperature $t_{sp}$ to soften or sharpen the probability distribution to ensure that the approximation is directly differentiable. Compared with the typical non-sparse approximation depicted in Eq.(8), Eq.(9) approximates $categorical(O)$ better by leveraging small temperature $t_{sp}$ to sparsify the output distribution of softmax (lower entropy). At the same time, Eq.(9) avoids the instability of training the supernet incurred by reparameterization in GDAS. From here, we refer $softmax ({\raise0.7ex\hbox{$A$} \!\mathord{\left/
 {\vphantom {A t}}\right.\kern-\nulldelimiterspace}
\!\lower0.7ex\hbox{$t_{sp}$}})$ with small temperature $t_{sp}$ as \textbf{sp}arse softmax and $softmax ({\raise0.7ex\hbox{$A$} \!\mathord{\left/
 {\vphantom {A t}}\right.\kern-\nulldelimiterspace}
\!\lower0.7ex\hbox{$t_{sm}$}})$ with normal temperature $t_{sm}$ as \textbf{sm}ooth softmax.

Softmax normalizes the input vector ${\bm{x}} = \{ {x_1},...,{x_d}\}$ to a probability distribution ${\bm{y}} = \{ {y_1},...,{y_d}\}$ where each entry can be calculated by ${y_i} = {\raise0.7ex\hbox{${\exp ({{{x_i}} \mathord{\left/
 {\vphantom {{{x_i}} t}} \right.
 \kern-\nulldelimiterspace} t})}$} \!\mathord{\left/
 {\vphantom {{\exp ({{{x_i}} \mathord{\left/
 {\vphantom {{{x_i}} t}} \right.
 \kern-\nulldelimiterspace} t})} {\sum\nolimits_{j = 1}^d {\exp ({{{x_j}} \mathord{\left/
 {\vphantom {{{x_j}} t}} \right.
 \kern-\nulldelimiterspace} t}))} }}}\right.\kern-\nulldelimiterspace}
\!\lower0.7ex\hbox{${\sum\nolimits_{j = 1}^d {\exp ({{{x_j}} \mathord{\left/
 {\vphantom {{{x_j}} t}} \right.
 \kern-\nulldelimiterspace} t})} }$}}$, Jacobian matrix of softmax is
\begin{equation}
\frac{{\partial {\bm{y}}}}{{\partial {\bm{x}}}} = \frac{1}{t}\left[ {\begin{array}{*{20}{c}}
{{y_1} - y_1^2}&{ - {y_1}{y_2}}&{ - {y_1}{y_3}}& \cdots &{ - {y_1}{y_d}}\\
{ - {y_2}{y_1}}&{{y_2} - y_2^2}&{ - {y_2}{y_3}}& \cdots &{ - {y_2}{y_d}}\\
 \vdots & \vdots & \vdots & \cdots & \vdots \\
{ - {y_d}{y_1}}&{ - {y_d}{y_2}}&{ - {y_d}{y_3}}& \cdots &{{y_d} - y_d^2}
\end{array}} \right]
\end{equation}
One noticeable shortcoming of the sparse softmax approximation depicted in Eq.(9) is that the gradient saturation occurs when one entry within the output is close to 1 and others are thus close to 0. When softmax is saturated, all entries of the Jacobian matrix depicted in Eq.(10) are close to 0 as $\mathop {\lim }\limits_{\scriptstyle{y_i} \to 1\hfill\atop
\scriptstyle{y_{j \ne i}} \to 0\hfill} \frac{{\partial {\bm{y}}}}{{\partial {\bm{x}}}} = 0$ and the backpropagation thereby stops propagating gradients through the softmax which is similar to the saturation of sigmoid. \textit{\textbf{We circumvent the gradient saturation by not solely considering the categorical approximation but jointly optimizing a combination of the approximations for both Eq.(5) and Eq.(7)}}. Concretely, we first combine Eq.(5) and Eq.(7) as shown in Eq.(11).
\begin{equation}
\mathop {\min }\limits_{\alpha,c} {L_{val}}(\omega _\alpha^ * (\bm{a}),\bm{a}) + {L_{val}}(\omega _c^ * (c),c)
\end{equation}
\begin{equation}
{\rm{s}}{\rm{.t}}{\rm{. }}\left\{ {\begin{array}{*{20}{c}}
\begin{aligned}
\omega _\alpha^ * (\bm{a}) &= \arg \mathop {\min }\limits_\omega  {L_{train}}(\omega ,\bm{a})\\
\omega _c^ * (c) &= \arg \mathop {\min }\limits_\omega  {L_{train}}(\omega ,c)\\
c &= categorical(O)\\
\bm{a} &= softmax (A)
\end{aligned}
\end{array}} \right.
\end{equation}
Then, we simultaneously approximate $categorical(O)$ and $softmax(A)$ by the sparse softmax ${{\bm{a}_{sp}} = softmax ({\raise0.7ex\hbox{$A$} \!\mathord{\left/
 {\vphantom {A {{t_{sp}}}}}\right.\kern-\nulldelimiterspace}
\!\lower0.7ex\hbox{${{t_{sp}}}$}})}$ and smooth softmax ${{\bm{a}_{sm}} = softmax ({\raise0.7ex\hbox{$A$} \!\mathord{\left/
 {\vphantom {A {{t_{sm}}}}}\right.\kern-\nulldelimiterspace}
\!\lower0.7ex\hbox{${{t_{sm}}}$}})}$ respectively in Eq.(13).
\begin{equation}
\mathop {\min }\limits_\alpha  {L_{val}}(\omega _\alpha^ * ({\bm{a}_{sm}}),{\bm{a}_{sm}}) + {L_{val}}(\omega _\alpha^ * ({\bm{a}_{sp}}),{\bm{a}_{sp}}){\rm{ }}
\end{equation}
\begin{equation}
{\rm{s}}{\rm{.t}}{\rm{.}}\left\{ {\begin{array}{*{20}{c}}
\begin{aligned}
&\omega _\alpha^ * ({\bm{a}_{sm}}) = \arg \mathop {\min }\limits_\omega  {L_{train}}(\omega ,{\bm{a}_{sm}})\\
&\omega _\alpha^ * ({\bm{a}_{sp}}) = \arg \mathop {\min }\limits_\omega  {L_{train}}(\omega ,{\bm{a}_{sp}})\\
&{\bm{a}_{sp}} = softmax ({\raise0.7ex\hbox{$A$} \!\mathord{\left/
 {\vphantom {A {{t_{sp}}}}}\right.\kern-\nulldelimiterspace}
\!\lower0.7ex\hbox{${{t_{sp}}}$}}){\rm{\ \ for\ }}{t_{sp}} = \frac{1}{{{{10}^n}}}{\rm{,\ {10}^n}} \gg {\rm{1}}\\
&{\bm{a}_{sm}} = softmax ({\raise0.7ex\hbox{$A$} \!\mathord{\left/
 {\vphantom {A {{t_{sm}}}}}\right.\kern-\nulldelimiterspace}
\!\lower0.7ex\hbox{${{t_{sm}}}$}}){\rm{\ \ for\ }}{t_{sm}} \gg {t_{sp}}
\end{aligned}
\end{array}} \right.{\rm{ }}
\end{equation}
where $\arg\mathop {\max }\limits_o (softmax ({\raise0.7ex\hbox{$A$} \!\mathord{\left/
 {\vphantom {A t}}\right.\kern-\nulldelimiterspace}
\!\lower0.7ex\hbox{$t$}}))$ does not change when we rescale the temperature of softmax, the ultimate goals of the two minimization objectives in Eq.(13) are thereby consistent. When ${t_{sm}} = 1$, the smooth softmax ${\bm{a}_{sm}}$ in Eq.(14) reduces to exactly the differentiable approximation $\bm{a}_{sm} = \bm{a}=softmax (A)$ of DARTS depicted in Eq.(8). Meanwhile, the second term $\mathop {\min }\limits_\alpha  {L_{val}}(\omega _\alpha^ * ({\bm{a}_{sp}}),{\bm{a}_{sp}}){\rm{ }}$ in Eq.(13) together with the sparse approximation ${\bm{a}_{sp}} = softmax ({\raise0.7ex\hbox{$A$} \!\mathord{\left/
 {\vphantom {A {{t_{sp}}}}}\right.\kern-\nulldelimiterspace}
\!\lower0.7ex\hbox{${{t_{sp}}}$}})$ in Eq.(14) can overall be regarded as an explicit sparse regularization with respect to the first term $\mathop {\min }\limits_\alpha  {L_{val}}(\omega _\alpha^ * (\bm{a}_{sm}),\bm{a}_{sm}){\rm{ }}$ and the smooth approximation $\bm{a}_{sm} = softmax (A)$ of DARTS. To alleviate the architecture mismatch, we aim to jointly optimize both the sparse and smooth approximation depicted in Eq.(13) to drive the supernet in the training phase closer to the discretized finalnet in the evaluation phase. We didn’t add an explicit coefficient, e.g. $\lambda $, in Eq.(13) to explicitly control the strength of the sparse regularization. In the next section, we exhibit that the strength can be implicitly controlled through a mixture coefficient of the batch sparsity. We call DARTS enhanced by our sparse regularization as Sparse DARTS or SP-DARTS.

\section{Batch-mixed sparse training}
Directly optimizing the proposed Eq.(13) normally needs to feedforward and backpropagate twice with different temperatures ${t_{sm}}$, ${t_{sp}}$ respectively and accumulate the gradients which inevitably doubles the computational overhead. To tackle this problem, we further devise a new training scheme by mixing batches across different temperatures within each training epoch to simultaneously minimize two objective terms in Eq.(13) in a joint manner. We use the mixed sparsity and mixed temperatures interchangeably in this section. 

\begin{figure}[ht]
\begin{center}
\includegraphics[width=\columnwidth]{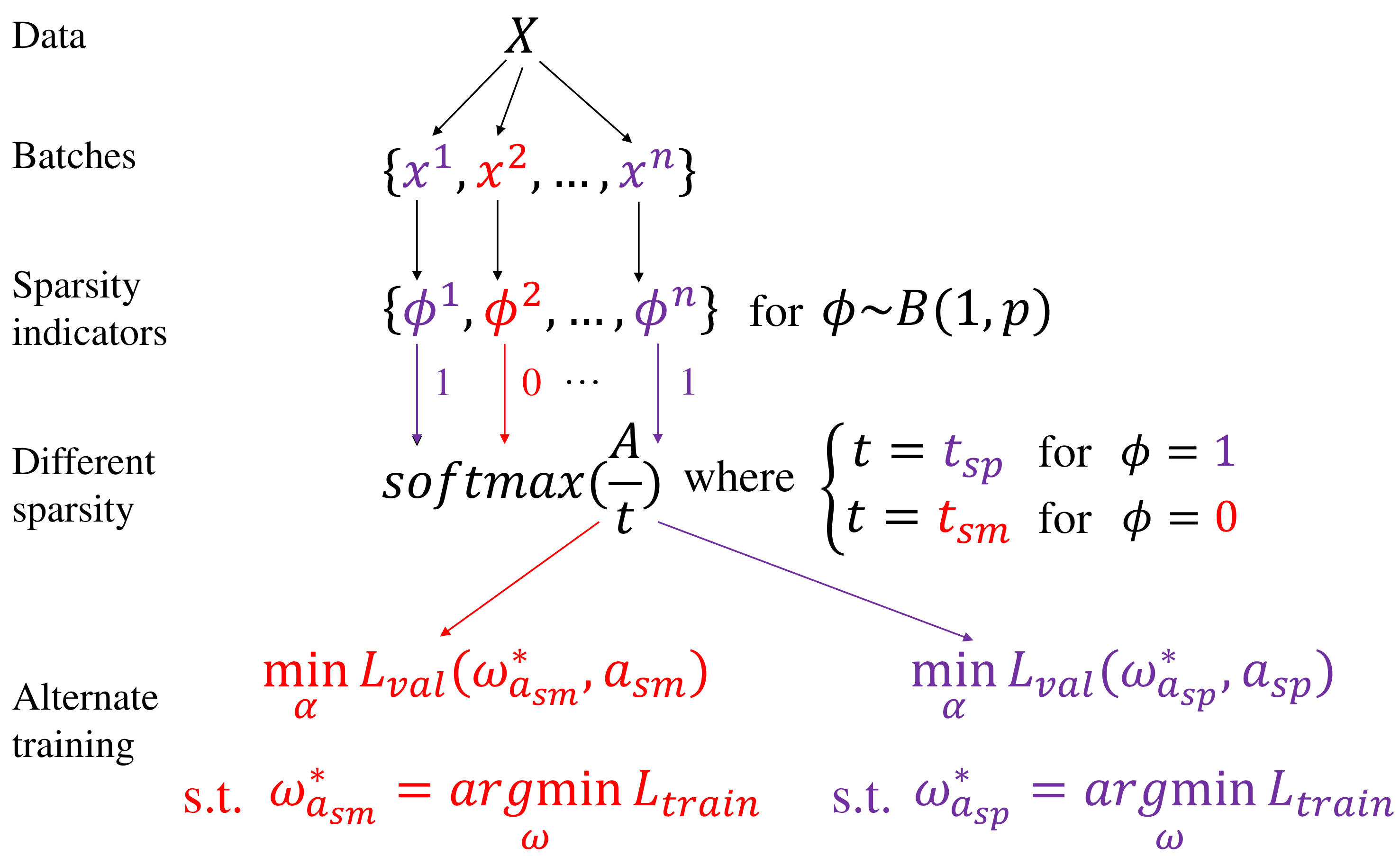}
\caption{Batches with different sparsity are trained alternately based on indicators ${\phi^i} \sim B(1,p)$.}
\label{fig2}
\end{center}
\end{figure}

All batch in this paper refers to the mini batch in SGD. Data $X$ is first divided into $N$ batches $X = \{ {x^1},{x^2},...,{x^N}\}$. We next introduce $\Phi  = \left\{ {{\phi^1},{\phi^2},...,{\phi^N}} \right\}$ to indicate the softmax temperature ${t^i} \in \{ {t_{sm}},{t_{sp}}\} $ corresponding to the batch ${x^i}$. The sparsity of the softmax output correlated with the indicator $\phi $ can be depicted in Eq.(15).
\begin{equation}
\bm{a}(\phi ) = \left\{ {\begin{array}{*{20}{c}}
{softmax ({\raise0.7ex\hbox{$A$} \!\mathord{\left/
 {\vphantom {A {{t_{sp}}}}}\right.\kern-\nulldelimiterspace}
\!\lower0.7ex\hbox{${{t_{sp}}}$}}){\rm{\ \ for }}\ \phi  = 1}\\
{softmax ({\raise0.7ex\hbox{$A$} \!\mathord{\left/
 {\vphantom {A {{t_{sm}}}}}\right.\kern-\nulldelimiterspace}
\!\lower0.7ex\hbox{${{t_{sm}}}$}}){\rm{\ \ for }}\ \phi  = 0}
\end{array}} \right.
\end{equation}
where we sample ${\phi^i} \sim B(1,p)$ and define $p$ as the probability of optimizing weights and parameters for a batch of data with temperature ${t_{sp}}$ depict in Eq.(16).
\begin{equation}
\mathop {\min }\limits_\alpha {L_{val}}(\omega _\alpha^ * ({\bm{a}_{sp}}),{\bm{a}_{sp}}){\rm{ }}
\end{equation}
\begin{equation}
{\rm{s}}{\rm{.t}}{\rm{.}}\left\{ {\begin{array}{*{20}{c}}
\begin{aligned}
&\omega _\alpha^ * ({\bm{a}_{sp}}) = \arg \mathop {\min }\limits_\omega  {L_{train}}(\omega ,{\bm{a}_{sp}})\\
&{\bm{a}_{sp}} = softmax ({\raise0.7ex\hbox{$A$} \!\mathord{\left/
 {\vphantom {A {{t_{sp}}}}}\right.\kern-\nulldelimiterspace}
\!\lower0.7ex\hbox{${{t_{sp}}}$}}){\rm{\ \ for\ }}{t_{sp}} = \frac{1}{{{{10}^n}}}{\rm{,\ {10}^n}} \gg {\rm{1}}
\end{aligned}
\end{array}} \right.{\rm{ }}
\end{equation}
On the contrary, $1-p$ is the probability of training the supernet based on the smooth approximation for a batch of data depicted in Eq.(18).
\begin{equation}
\mathop {\min }\limits_\alpha  {L_{val}}(\omega _\alpha^ * ({\bm{a}_{sm}}),{\bm{a}_{sm}})
\end{equation}
\begin{equation}
{\rm{s}}{\rm{.t}}{\rm{.}}\left\{ {\begin{array}{*{20}{c}}
\begin{aligned}
&\omega _\alpha^ * ({\bm{a}_{sm}}) = \arg \mathop {\min }\limits_\omega  {L_{train}}(\omega ,{\bm{a}_{sm}})\\
&{\bm{a}_{sm}} = softmax ({\raise0.7ex\hbox{$A$} \!\mathord{\left/
 {\vphantom {A {{t_{sm}}}}}\right.\kern-\nulldelimiterspace}
\!\lower0.7ex\hbox{${{t_{sm}}}$}}){\rm{\ \ for\ }}{t_{sm}} \gg {t_{sp}}
\end{aligned}
\end{array}} \right.{\rm{ }}
\end{equation}
In this way, the coefficient $p$ within $B(1,p)$ acts as a mixture probability determines the ratio of the different sparsity batches within each epoch which can be utilized to indirectly controls the strength of the sparse regularization depicted in Eq.(13).
By sampling from the Bernoulli distribution, we also introduce random noise to the supernet training which has been shown to be beneficial for DARTS by Smooth DARTS \cite{chen2020stabilizing}. Figure~\ref{fig2} visually illustrates the process of alternately optimizing the first and second term of the minimization objective in Eq.(13) over different-sparsity batches.

\begin{table*}[t]
\caption{Experimental results on NAS-BENCH-201 on CIFAR-10 (first group) and CIFAR-100 (second group).}
\label{table1}
\begin{center}
\begin{tabular}{cccccccc}
\toprule
\multirow{2}{*}{Method} & \multirow{2}{*}{\begin{tabular}[c]{@{}c@{}}Search\\ (seconds)\end{tabular}} & \multicolumn{2}{c}{CIFAR-10} & \multicolumn{2}{c}{CIFAR-100} & \multicolumn{2}{c}{ImageNet-16-120} \\
 &  & validation & test & validation & test & validation & test \\
\midrule
DARTS-V2 \cite{liu2018darts} & 22323 & 39.77±0.00 & 54.30±0.00 & 15.03±0.00 & 15.61±0.00 & 16.43±0.00 & 16.32±0.00 \\
DARTS-V1 \cite{liu2018darts} & 7253 & 39.77±0.00 & 54.30±0.00 & 15.03±0.00 & 15.61±0.00 & 16.43±0.00 & 16.32±0.00 \\
GDAS \cite{dong2019searching} & 19720 & 90.00±0.21 & 93.51±0.13 & 71.14±0.27 & 70.61±0.26 & 41.70±1.26 & 41.84±0.90 \\
DrNAS \cite{chen2021drnas} & 7544 & 90.15±0.10 & 93.74±0.03 & 70.82±0.27 & 71.07±0.08 & 40.76±0.05 & 41.37±0.17 \\
GAEA-bilevel \cite{li2021geometry} & 8280 & 82.80±1.01 & 84.64±1.00 & 55.24±1.47 & 55.35±1.72 & 27.72±1.35 & 26.40±0.85 \\
GAEA-ERM \cite{li2021geometry} & 14464 & 84.59±0.00 & 86.59±0.00 & 58.12±0.00 & 58.43±0.00 & 29.54±0.00 & 28.19±0.00 \\
 &  & (91.50±0.06) & (94.34±0.06) & (73.12±0.26) & (73.11±0.06) & (45.71±0.28) & (46.38±0.18) \\
\textbf{SP-DARTS} & 7316 & \textbf{91.25±0.23} & \textbf{94.01±0.31} & \textbf{72.11±1.04} & \textbf{72.31±0.67} & \textbf{45.16±0.89} & \textbf{45.26±0.92} \\
\midrule
PC-DARTS \cite{xu2019pc} & - & 89.96±0.15 & 93.41±0.30 & 67.12±0.39 & 67.48±0.89 & 40.83±0.08 & 41.31±0.22 \\
GDAS \cite{dong2019searching} & - & 90.01±0.46 & 93.23±0.23 & 24.05±8.12 & 24.20±8.08 & 40.66±0.00 & 41.02±0.00 \\
DrNAS \cite{chen2021drnas} & - & 90.00±0.12 & 93.69±0.04 & 71.20±0.30 & 70.96±0.09 & 40.69±0.05 & 41.12±0.19 \\
GAEA-bilevel \cite{li2021geometry} & - & 82.45±0.16 & 84.16±0.01 & 54.64±0.19 & 54.68±0.13 & 27.13±0.10 & 26.12±0.06 \\
GAEA-ERM \cite{li2021geometry} & - & 76.49±6.70 & 79.82±5.60 & 51.19±5.74 & 51.18±6.00 & 21.70±6.49 & 20.51±6.35 \\
 &  & (90.06±0.47) & (93.33±0.00) & (71.56±0.10) & (71.22±0.86) & (42.30±0.24) & (42.28±0.13) \\
\textbf{SP-DARTS} & - & \textbf{91.23±0.19} & \textbf{94.02±0.20} & \textbf{71.71±1.08} & \textbf{71.85±0.90} & \textbf{45.15±0.93} & \textbf{45.30±0.72} \\
\midrule
ResNet & N/A & 90.83 & 93.97 & 70.42 & 70.86 & 44.53 & 43.63 \\
Optimal & N/A & 91.61 & 94.37 & 73.49 & 73.51 & 46.77 & 47.31\\
\bottomrule
\end{tabular}
\end{center}
\caption{SP-DARTS experimental configurations on NAS-BENCH-201.}
\label{table2}
\begin{center}
\begin{tabular}{cccccccccc}
\toprule
Dataset & Parameter\_Decay & Weight\_LR & Eta\_min & ${p_{low}}$ & ${p_{up}}$ & ${t_{sp}}$ & ${t_{sm}}$ & ${e_{wp}}$ & Batch\_size \\
\midrule
CIFAR-10/100 & 0 & 0.05 & 0.05 & 0 & 1/0.9 & 1.1e-3/1.3e-3 & $10\times{t_{sp}}$ & 1 & 256 \\
\bottomrule
\end{tabular}
\end{center}
\end{table*}

Initialization of the supernet always has non-trivial impact on the optimization of DARTS thus empirically, a predefined fixed $p$ shows insufficient robustness. We schedule the value of $p$ with respect to the training epoch $i$ as well as predefined upper and lower bounds ${p_{up}}$, ${p_{low}}$ during training. Algorithm 1 exhibits the complete batch-mixed sparse training scheme.
\begin{algorithm}[tb]  
   \caption{Batch-Mixed Sparse Training Scheme}
   \label{alg:1}
\begin{algorithmic}
   \STATE {\bfseries Input:} Upper bound ${p_{up}}$ and lower bound ${p_{low}}$, temperature ${t_{sm}}$ and ${t_{sp}}$, total number of epochs $I$, batch size $b$, training data ${X_{train}}$, validation data ${X_{val}}$, warmup epochs ${i_{wp}}$.
   \STATE $N = {\raise0.7ex\hbox{${len({X_{train}})}$} \!\mathord{\left/
 {\vphantom {{len({X_{train}})} b}}\right.\kern-\nulldelimiterspace}
\!\lower0.7ex\hbox{$b$}}$
   \WHILE{training epoch $i \le I$}
   \STATE ${p_i} = {P_{schedule}}({p_{up}},{p_{low}},I,i,{i_{wp}})$
   \STATE $\Phi  = \left\{ {{\phi ^1},{\phi ^2},...,{\phi ^n}} \right\}$ for $\phi \sim B(1,{p_i})$
   \WHILE{batches $x_{train}^n \in \{ x_{train}^1,x_{train}^2,...,x_{train}^N\} $ and $x_{val}^n \in \{ x_{val}^1,x_{val}^2,...,x_{val}^N\} $}
   \STATE Update weights $\omega $ with ${\nabla _\omega }{L_{train}}(\omega ,\bm{a}({\phi ^n}),x_{train}^n)$
   \STATE Update parameters $\alpha $ with ${\nabla _\alpha }{L_{val}}(\omega ,\bm{a}({\phi ^n}),x_{val}^n)$
   \ENDWHILE
   \ENDWHILE
\end{algorithmic}
\end{algorithm}
We step forward to formulate the scheduler for $p$ in Eq.(20).
\begin{equation}
\begin{array}{l}
{P_{schedule}}({p_{up}},{p_{low}},I,i,{i_{wp}})\\
 = \left\{ \begin{array}{*{20}{c}}
\begin{aligned}
&{\rm{0\ \ for\ }}i < {i_{wp}}&\\
&{p_{low}} + \frac{{{p_{up}} - {p_{low}}}}{{I - {i_{wp}}}}(i - {i_{wp}}){\rm{\ \ for\ }}{i_{wp}} \le i \le I
\end{aligned}
\end{array} \right.
\end{array}
\end{equation}
where $p$ is linearly boosted from ${p_{low}}$ to ${p_{up}}$ to gradually enhance the strength of the sparse regularization with respect to the training epoch $i$ after warming up ${i_{wp}}$ epochs. This way, we can efficiently achieve sparse regularization by mixing batches across different sparsity within each epoch instead of feedforward and backpropagation twice for each batch of data.

\section{Experiments}
Our source code is online available\footnote{https://github.com/chaoji90/SP-DARTS}. In this section, we evaluate the performance of SP-DARTS on four datasets (CIFAR-10, CIFAR-100, ImageNet, SVHN) and multiple search spaces (NAS-BENCH-201 \cite{dong2019bench}, DARTS search space \cite{liu2018darts}, S1$\sim$S4 from \cite{arber2020understanding}).

\subsection{Evaluations on NAS-BENCH-201}
NAS-BENCH-201 supports three datasets (CIFAR-10, CIFAR-100, ImageNet-16-120) and has a unified cell-based search space with 15,625 architectures. We notice some deviations from the experimental configurations employed by the original benchmark \cite{dong2019bench} across literature. For fair comparison, we first declare the search space (five operations: none, skip connection, 1$\times$1 convolution, 3$\times$3 convolution, average pool) and the amount of training epochs (50) corresponding to the code released by NAS-BENCH-201\footnote{https://github.com/D-X-Y/NAS-Bench-201} as standard 201. All our experiments are based on the standard 201 for fair comparison to the benchmarks from the original paper. We refer to their paper \cite{dong2019bench} for more details of the search space. As with \cite{dong2019bench}, throughout, we report the average accuracies and standard deviations by searching under three different seeds.

\begin{table*}[t]
\caption{Experimental results on DARTS search space.}
\label{table3}
\begin{center}
\begin{tabular}{cccccccc}
\toprule
\multirow{2}{*}{Architecture} & \multicolumn{2}{c}{CIFAR-10} & \multicolumn{3}{c}{ImageNet (mobile settings)} & \multirow{2}{*}{\begin{tabular}[c]{@{}c@{}}Search Cost \\ (GPU days)\end{tabular}} & \multirow{2}{*}{Search Method} \\
 & Test   Error (\%) & Params   (M) & Top-1   (\%) & Top-5   (\%) & Params   (M) &  &  \\
\midrule
DenseNet-BC   (no cutout) & 3.46 & 25.6 & - & - & - & - & manual \\
\midrule
NASNet-A & 2.65 & 3.3 & 26.0 & 8.4 & 5.3 & 2000 & RL \\
AmoebaNet-B & 2.55   ± 0.05 & 2.8 & 26.0 & 8.5 & 5.3 & 3150 & evolution \\
PNAS   (no cutout) & 3.41   ± 0.09 & 3.2 & 25.8 & 8.1 & 5.1 & 225 & SMBO \\
ENAS \cite{pham2018efficient}& 2.89 & 4.6 & - & - & - & 0.5 & RL \\
\midrule
DARTS-V2 \cite{liu2018darts}& 2.76  ± 0.09 & 3.3 & 26.7 & 8.7 & 4.7 & on   CIFAR-10 & Gradient \\
PC-DARTS \cite{xu2019pc}& 2.57 ± 0.07 & 3.6 & 25.1 & 7.8 & 5.3 & on   CIFAR-10 & Gradient \\
GDAS \cite{dong2019searching}& 2.85 ± 0.02 & 2.8 & 26.0 & 8.5 & 5.3 & on   CIFAR-10 & Gradient \\
GAEA-PC-DARTS \cite{li2021geometry}& 2.50 ± 0.06 & 3.7 & 24.3 & 7.3 & 5.6 & on   CIFAR-10 & Gradient \\
DrNAS \cite{chen2021drnas}& 2.54 ± 0.03/2.46 ± 0.03 & 4.0/4.1 & 24.2/23.7 & 7.3/7.1 & 5.2/5.7 & on   ImageNet & Gradient \\
DARTS+PT \cite{wang2020rethinking}& 2.61  ± 0.08 & 3.0 & - & - & - & - & Gradient \\
SDARTS-RS+PT \cite{wang2020rethinking}& 2.54 ± 0.10 & 3.3 & - & - & - & - & Gradient \\
SGAS+PT \cite{wang2020rethinking}& 2.56 ± 0.10 & 3.9 & - & - & - & - & Gradient \\
\textbf{SP-DARTS} & 2.50 ±  0.07 & 3.5 & 24.5 & 7.6 & 4.9 & on   CIFAR-10 & Gradient \\
\bottomrule
\end{tabular}
\end{center}
\caption{SP-DARTS experimental configurations on DARTS search space.}
\label{table4}
\begin{center}
\begin{tabular}{cccccccccc}
\toprule
Dataset & Parameter\_ Decay & Weight\_LR & Eta\_min & ${p_{low}}$ & ${p_{up}}$ & ${t_{sp}}$ & ${t_{sm}}$ & ${e_{wp}}$ & Batch\_size \\
\midrule
CIFAR-10 & 0 & 0.05 & 0.05 & 0 & 1 & 1.0e-3 & $10\times{t_{sp}}$ & 1 & 80 (memory limit) \\
\bottomrule
\end{tabular}
\end{center}
\end{table*}

We introduce DrNAS \cite{chen2021drnas} and GAEA \cite{li2021geometry} from ICLR 2021 as two baselines of our evaluations on NAS-BENCH-201. Besides, GDAS \cite{dong2019searching} is the art method in the original paper of NAS-BENCH-201 and PC-DARTS \cite{xu2019pc} is also a well-developed approximate SOTA model. We provide both the searching results on CIFAR-10 and CIFAR-100 in the first and second group respectively in Table~\ref{table1}.

Table~\ref{table2} lists the detailed experimental configurations of the training phase on NAS-BENCH-201. CIFAR-100 has 10$\times$ more labeled categories than CIFAR-10, resulting in higher gradient variances. We slightly increase and temperature parameter ${t_{sp}}$ to stabilize the training on CIFAR-100. Generally, we would like to set ${t_{sp}}$ to be a relative small value and tune ${p_{up}}$ to control the regularization strength.
All unlisted settings are consistent with the default configurations of NAS-BENCH-201. We find that the GAEA-ERM is particularly fragile for the “none” operation on standard 201. Therefore, we also provide additional results for GAEA-ERM alone in parenthesis by excluding the none operation in the search space. Experiments of DrNAS\footnote{https://github.com/xiangning-chen/DrNAS} and GAEA\footnote{https://github.com/liamcli/gaea_release} are both based on the source codes released by their authors. Note that when comparing with prior art baselines, SP-DARTS claims a clearly preferable performance on standard 201 on CIFAR-100 in our experiment. In addition, SP-DARTS outperforms all other baselines on CIFAR-10 except for GAEA-ERM on a much smaller (only 4096 architectures) and less noisy (exclude none) search space.

DrNAS converges very fast and consistently end up with arch-index=1462 (six 3$\times$3 convolutions) or arch-index=138 (five 3$\times$3 convolutions and one none operation) even under different seeds on different datasets. It’s a little strange that an one-shot method always converges to the same architecture since the gradients within neural network optimization are notoriously noisy and greatly affected by initialization. Oracle optimal in the search space refers to the average performance rather than the consistent score under all different seeds \cite{yang2019evaluation}. Even further, the accurate rankings of the top performing architectures are usually only discernible after several hundred epochs of training. It seems impractical to expect that the one-shot method can consistently get the optimal performance under all different seeds within only 50 epochs of training. If the one-shot method always converges to the same architecture, we suspect that it is due to a strong prior and doubt whether the method is really searching in this case.

We provide a comparison of the relative time cost between SP-DARTS and other baselines under the same training epoch (50) on standard 201 on CIFAR-10 in the second column of the first group in Table~\ref{table1}. By investigating the source code, we also ensure the same following conditions: 
\begin{itemize}
\setlength{\itemsep}{0pt}
\setlength{\parsep}{0pt}
\setlength{\parskip}{0pt}
\item Environments: Same gpu. Same software version; 
\item Settings: Batch size (160). Init channel scale (24); 
\item Implementations: Do not query the performance database. Do not evaluate supernet with the test set every epoch. Save checkpoint every epoch.
\end{itemize}
We find that the time cost of SP-DARTS is among the lowest level of all baselines which is quite close to DARTS-V1. This demonstrates the efficiency of our batch-mixed training scheme. The time costs of GDAS and DARTS-V2 are normalized based on the time cost of DARTS-V1 in our experiment and the results reported by the original NAS-BENCH-201 \cite{dong2019bench}.

\begin{table*}[!h]
\caption{Experimental results on S1$\sim$S4 search spaces.}
\label{table5}
\begin{center}
\resizebox{\textwidth}{!}{
\begin{tabular}{ccccccccc}
\toprule
Dataset & Space & DARTS & PC-DARTS & DARTS-ES & R-DARTS(DP/L2) & SDARTS(RS/ADV) & \begin{tabular}[c]{@{}c@{}}DARTS+PT\\    (unfixed/fixed) \end{tabular} & \textbf{SP-DARTS} \\
\midrule
\multirow{4}{*}{CIFAR-10} & S1 & 3.84 & 3.11 & 3.01 & 3.11/2.78 & 2.78/2.73 & 3.50/2.86 & \textbf{2.70} \\
&S2 & 4.85 & 3.02 & 3.26 & 3.48/3.31 & 2.75/2.65 & 2.79/\textbf{2.59} & 2.66 \\
&S3 & 3.34 & 2.51 & 2.74 & 2.93/2.51 & 2.53/\textbf{2.49} & \textbf{2.49}/2.52 & 2.51 \\
&S4 & 7.20 & 3.02 & 3.71 & 3.58/3.56 & 2.93/2.87 & 2.64/\textbf{2.58} & 2.60 \\
\midrule
\multirow{4}{*}{CIFAR-100} & S1 & 29.46 & 24.69 & 28.37 & 25.93/24.25 & 23.51/22.33 & 24.48/24.40 & \textbf{22.30} \\
&S2 & 26.05 & 22.48 & 23.25 & 22.30/22.44 & 22.28/20.56 & 23.16/23.30 & \textbf{20.55} \\
&S3 & 28.90 & 21.69 & 23.73 & 22.36/23.99 & 21.09/21.08 & 22.03/21.94 & \textbf{21.04}\\
&S4 & 22.85 & 21.50 & 21.26 & 22.18/21.94 & 21.46/21.25 & 20.80/\textbf{20.66} & 21.49\\
\midrule
\multirow{4}{*}{SVHN} & S1 & 4.58 & 2.47 & 2.72 & 2.55/4.79 & 2.35/\textbf{2.29} & 2.62/2.39 & 2.33 \\ 
& S2 & 3.53 & 2.42 & 2.60 & 2.52/2.51 & 2.39/2.35 & 2.53/2.32 & \textbf{2.30} \\ 
& S3 & 3.41 & 2.41 & 2.50 & 2.49/2.48 & 2.36/2.40 & 2.42/\textbf{2.32} & \textbf{2.32} \\ 
& S4 & 3.05 & 2.43 & 2.51 & 2.61/2.50 & 2.46/2.42 & 2.42/\textbf{2.39} & 2.42\\
\bottomrule
\end{tabular}
}
\end{center}
\end{table*}

\subsection{Evaluations on DARTS search space}
DARTS search space excludes the none operation and is much larger than NAS-BENCH-201, both of which result in much more stable training of SP-DARTS. Most of the hyperparameter configurations can be directly transformed from NAS-BENCH-201 or with some slightly modifications shown in Table~\ref{table4}. All unlisted settings are consistent with the default configurations of DARTS \cite{liu2018darts}. We provide the experimental results on DARTS search space in Table~\ref{table3}. Our architecture evaluations are based on the source code released by DrNAS \cite{chen2021drnas}. We keep all hyperparameter settings unchanged except for replacing the cell genotypes. Most of our baselines \cite{chen2021drnas,li2021geometry,wang2020rethinking} from ICLR2021 are still very strong. We find that the existing SOTA methods always try to find large-scale architectures (more parameters), by contrast to our method which inclines to look for more efficient cell. As with \cite{li2021geometry}, we repeat SP-DARTS five times under different seeds. The best architecture achieves 97.6\% accuracy with only 3.4M parameters on CIFAR-10 in our experiment. 

As a common practice, we transfer the architecture from CIFAR-10 to Imagenet for additional performance evaluation. The results are shown in fourth, fifth and sixth columns of Table~\ref{table3}. Since the search time of the gradient-based methods are significantly affected by which dataset is used for training and it is always difficult to fairly normalize the time overhead from different papers, we only specify the search datasets corresponding to the relevant scores in the time cost column (seventh column) for all gradient-based methods in Table~\ref{table3}. 

\subsection{Comparison with Other Regularization}
Another important evaluation is conducted on four specially designed search spaces S1$\sim$S4 crafted by \cite{arber2020understanding} to further validate the effectiveness of our method. Unregularized DARTS is always prone to choose non-parametric operations on these search spaces. We refer to \cite{arber2020understanding} for more details about S1$\sim$S4. We evaluate SP-DARTS against two strong baselines Smooth DARTS \cite{chen2020stabilizing} and DARTS+PT \cite{wang2020rethinking} on this benchmark on CIFAR-10, CIFAR-100 and SVHN respectively. Most hyperparameter configurations in the training phase are consistent with the experiments on DARTS search space shown in Table~\ref{table4}. We do not systematically tune hyperparameters, but choose setups that work reasonably well across different datasets and search spaces.
According to \cite{arber2020understanding,chen2020stabilizing}, our results in Table~\ref{table5} are obtained by running SP-DARTS four times with different values of ${t_{sp}}$ (4e-4, 5e-4, 6e-4, 7e-4) on each search space for each dataset respectively and pick the finalnet based on the validation accuracy. Our architecture evaluations are based on the source code released by Smooth DARTS\footnote{https://github.com/xiangning-chen/SmoothDARTS}. Again, we keep the hyperparameter configurations unchanged except for replacing the cell genotypes. 

Table~\ref{table5} shows the results of the evaluation on S1$\sim$S4 search spaces. We need to point out that both SDARTS-ADV and DARTS+PT call for an extra step to employ fine-tuning or adversarial attack both of which incur considerably additional time overhead comparing with DARTS reported by their paper.
By employing our specific batch-mixed training scheme, SP-DARTAS achieves the performance on par with SDARTS-ADV \cite{chen2020stabilizing} and DARTS+PT \cite{wang2020rethinking} without the extra overhead. Otherwise, as shown in Table~\ref{table5}, SP-DARTS outperforms all other methods including DARTS \cite{liu2018darts}, R-DARTS(L2) \cite{arber2020understanding}, DARTS-ES \cite{arber2020understanding}, R-DARTS(DP) \cite{arber2020understanding}, and PC-DARTS \cite{xu2019pc}.

\begin{table*}[t]
\caption{Accuracies of DARTS with 0.1 parameter learning rate on NAS-BENCH-201 on CIFAR-10 and CIFAR-100.}
\label{table6}
\begin{center}
\begin{tabular}{ccccccc}
\toprule
\multirow{2}{*}{Method} & \multicolumn{2}{c}{CIFAR-10} & \multicolumn{2}{c}{CIFAR-100} & \multicolumn{2}{c}{ImageNet-16-120} \\
 &  validation & test & validation & test & validation & test \\
\midrule
DARTS-0.1 & 84.04±1.84 (-7.21) & 86.43±2.21 (-7.58)& 58.42±5.17 (-13.69) & 58.39±4.71 (-13.92) & 30.08±3.51 (-15.08) & 29.19±4.24 (-16.07)\\
\midrule  
DARTS-0.1 & 85.24±1.93 (-5.99) & 87.36±2.21 (-6.66)& 59.31±3.69 (-12.4) & 59.69±3.73 (-12.16) & 31.64±4.10 (-13.51) & 29.85±4.67 (-15.45) \\
\bottomrule
\end{tabular}
\end{center}
\end{table*}

\subsection{Analysis and Discussion}
\begin{figure}[t]
\begin{center}
\includegraphics[width=\columnwidth]{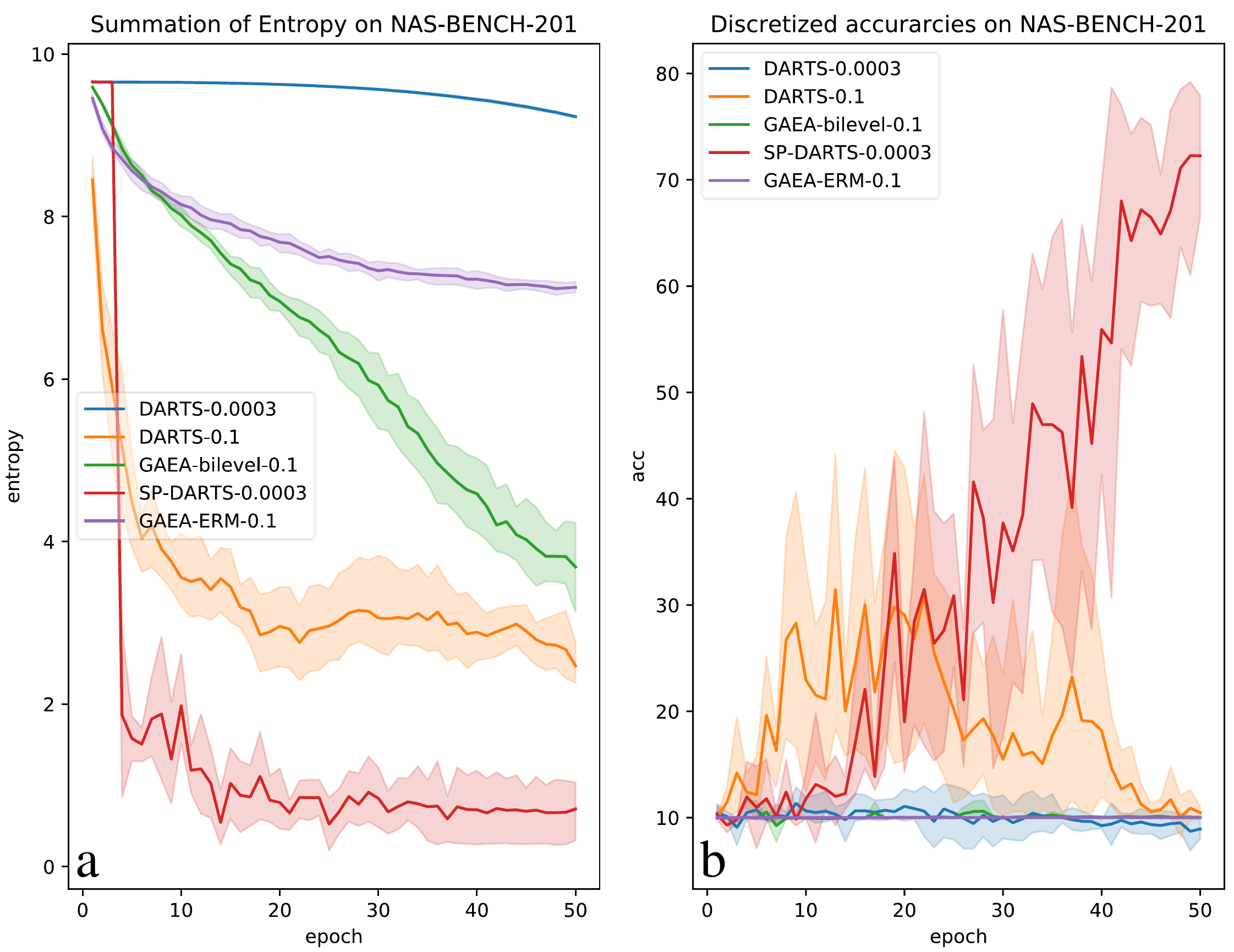}
\caption{(a) IES values in the training phases. (b) Discretized accuracies on validation set in the training phases.}
\label{fig3}
\end{center}
\end{figure}

As \cite{li2021geometry}, we assess the sparsity of parameter distribution in the training phase by introducing the information entropy summation (IES) over compound edges depicted in Eq.(21).
\begin{equation}
IES =  - \sum\limits_{e=1}^E {\sum\limits_{m=1}^M {a_e^m\log a_e^m} }\ {\rm{for}}\ {\bm{a}_{e}} = softmax ({A_{e}})
\end{equation}
where $E$ compound edges and $M$ candidate operations in the search space. $a_e^m$ denotes the entry of the operation $m$ within parameter distribution $\bm{a}_{e}$ on edge $e$. IES calculates the information entropy across all operation candidates within each compound edge and sums them up over all edges in the search space. Figure~\ref{fig3}(a) presents the entropy values over the training epochs on NAS-BENCH-201 on CIFAR-10 from which we can clearly recognize that SP-DARTS achieves significantly sparser parameters (lower entropy) than all other baselines. Evident by Table~\ref{table1} and Table~\ref{table6}, we can easily notice the performance improvement of DARTS-0.1 compared to DARTS-0.0003 and the performance gap compared to SP-DARTS.
DARTS-0.1 coarsely enhances the parameter sparsity by increasing the parameter learning rate (0.0003$\to$0.1), but it is still helpful to alleviate the performance collapse shown in Table~\ref{table6} which clearly emphasizes the regularization effect of the sparse training. However DARTS-0.1 encounters a bottleneck when the IES reaches about 3.5 which dampens the sparse training and results in insufficient sparsity to suppress information bypass leakage. 
The impact of this leakage can be seen more clearly by the discretized accuracies shown in Figure~\ref{fig3}(b). 

Figure~\ref{fig3}(b) illustrates the supernet discretized accuracies according to the epochs in the training phase. The discretized architectures are extracted by Eq.(4). We can see in Figure~\ref{fig3}(b) that the DARTS-0.0003 always remains at the random level of accuracy under the smooth approximation which verifies our suspicion that the supernet actually learns the coadapated weights between operations. In this case, the magnitudes of parameters are easily failing to reflect the importance of operations. By increasing the parameter learning rate from 0.0003 to 0.1, the discretized accuracy of DARTS-0.1 is improved in the middle phase of training as shown by the orange line in Figure~\ref{fig3}(b). However, when the sparsity of parameter is dampened by saturation, DARTS-0.1 fails to match the discretized finalnet but converge to the random accuracy at the end of training. This clearly demonstrates that the parameter sparsity caused by increasing the parameter learning rate from 0.0003 to 0.1 is not enough to suppress information bypass leakage, so it’s incompetent to prevent the performance inconsistency between supernet and finalnet. SP-DARTS in contrast shown by the red line in Figure~\ref{fig3}(b), gradually increases the intensity of the sparse regularization under our training scheme by which SP-DARTS significantly improves the discretized accuracy as the supernet converges at the end of the training phase. 

Sparser parameters in SP-DARTS intrinsically make the continuous relaxation closer to the categorical selection which naturally breaks the advantage of the skip connection because the predictive information can no longer be transmitted through bypasses. Sparse training thereafter forces supernet to directly attribute stronger strength to more predictive operations in terms of the magnitudes of parameters. Beyond that, the sparser parameters lead to supernet significantly more similar to the discretized finalnet in the training phase which again inherently alleviate architecture mismatch in the post-search discretization step and is thereby vital to the validity of both the underlying hypotheses 1 and 2 of DARTS.

\section{Related Works}
Broadly speaking, NAS can be abstracted as a combinatorial problem. Black-box optimization methods to deal with this problem have been comprehensively studied for a long time, including Bayesian optimization, evolutionary algorithm, reinforcement learning, etc. In theory, these algorithms can be used almost literally in NAS, thus the NAS idea itself is nothing innovative. 

However, NAS research did not appear on the mainstream of the deep learning community until 2017 \cite{zoph2016neural,zoph2018learning} which is mainly due to two reasons: 1. Black-box optimization methods are mostly known for their sample insensitivity and inefficiency; 2. The huge budget of evaluating the performance of neural networks. Superposition of both two factors leads to the prohibitively expensive computational cost when the traditional optimization methods are blindly applied to NAS. Nevertheless, the early NAS researches still started according to this way which typically took thousands of GPU hours made NAS experiments almost unaffordable. Even so, the most important contribution of the early researches was that they demonstrated, at great cost, that NAS could indeed obtain the architectures exceeded SOTA performances of the handcrafted neural networks. This achievement has a profound impact on both NAS itself and the whole deep learning community. On the one hand, many subsequent SOTA networks proposed by the deep learning community, especially involving CV tasks, referred more or less to the network architecture obtained by NAS \cite{real2019regularized,howard2019searching,tan2019efficientnet,tan2020efficientdet}. On the other hand, a lot of researches on NAS itself have turned to improve the search efficiency and reduce the computational budget.

Typical improvements include surrogate model, proxy dataset, using meta-learning to predict network performance, predicting network parameters by hypernet, etc. Among other things, the cell-based search paradigm is an important step based upon the observation that many effective handcrafted architectures were designed with repetitions of fixed well-designed structure. Cell-based search paradigm non-trivially reduce the search space and the result can be stacked to achieve flexible network capacity which enables easy generalization across datasets and tasks. NAS cell is usually abstracted as a directed acyclic graph (DAG) \cite{elsken2019neural,wistuba2019survey}, where the edges represent operations and the nodes represent feature maps that link operations together. Another important step was the proposal of one-shot NAS based on the supernet and weights sharing \cite{bender2018understanding,dong2019one}. Gradient-based NAS methods utilize the gradients obtained internally in the training process as the supervision signal, which is clearly different from the previous black-box optimization NAS. Gradient-based methods have quickly become the mainstream research direction due to their high efficiency. This direction was subsequently developed as DARTS \cite{liu2018darts,dong2019searching,xu2019pc} which explicitly parameterizes the architecture and optimizes it through the generic loss gradients from backpropagation in a bilevel optimization scheme. The performance of DARTS is therefore directly correlated with SGD optimization dynamics.

Traditional deep learning researches are often trapped in the reproducibility. NAS methods usually include two stages: search and evaluation, which involve more degrees of freedom and computational budget. As a result, NAS researches are even more difficult to reproduce than traditional deep learning researches \cite{yang2019evaluation}. Comparing performance between different NAS methods are still an open question. Some studies have tried to partially solve this problem by introducing NAS-BENCH \cite{dong2019bench,zela2019bench,ying2019bench} which primarily addresses the problem of reproducibility and computational budget of the NAS experiments. Nevertheless, NAS-BENCH itself also brings additional overfitting risk to the NAS methods.

\section{Conclusion}
In this paper, we aim to robustify DARTS by eliminating information bypass leakage. Specifically, we propose to simultaneously approximate categorical and continuous approximation by sparse and smooth softmax to alleviate the premature convergence caused by the softmax gradient saturation and achieve much sparser parameters than ever before in the training phase of DARTS. The sparse approximation term in the objective acts as an sparse regularization against the smooth approximation term in DARTS. 
We then propose the batch-mixed training scheme to achieve sparse training efficiently by mixing different sparsity mini batches within each training epoch. We implicitly control the strength of the regularization by adjusting a mixture coefficient and schedule it in the training phase so that the parameters gradually converge to a sparse solution which eventually match the architecture of the discretized finalnet. We subsequently conduct extensive experiments to verify the effectiveness of our method.

\section*{Acknowledgment}
The work is supported by National Natural Science Foundation of China (Grant No. 61703013 and No. 91646201) and National Key R\&D Program of China (No. 2017YFC0803300).

\bibliographystyle{IEEEtran}
\bibliography{icdm2021}

\end{document}